\documentclass{article}
\usepackage[preprint]{neurips2019}
\usepackage{times}
\usepackage{amsmath}
\usepackage{amssymb}
\usepackage{latexsym}
\usepackage{booktabs}
\usepackage{graphicx}
\usepackage{multirow}
\usepackage{dsfont}
\usepackage{float}
\usepackage{graphicx}
\usepackage{array}
\usepackage[utf8]{inputenc} 
\usepackage[T1]{fontenc}    
\usepackage{hyperref}       
\usepackage{url}            
\usepackage{booktabs}       
\usepackage{amsfonts}       
\usepackage{nicefrac}       
\usepackage{microtype}      
\usepackage{xcolor}
\usepackage{wrapfig}
\usepackage{multicol}
\usepackage{makecell}

\usepackage{url}
\setcitestyle{numbers,square}

\newcommand\modelname{Transformer-SM}
\newcommand\modelnameshort{T-SM}

\newcolumntype{L}{>{\arraybackslash}m{12cm}}

\title{Efficient Adaptation of Pretrained Transformers for Abstractive Summarization}

\author{Andrew Hoang$^{\heartsuit}$, Antoine Bosselut$^{\heartsuit\spadesuit}$, Asli Celikyilmaz$^\clubsuit$, Yejin Choi$^{\heartsuit\spadesuit}$\\
$^\heartsuit$ Allen School of Computer Science \& Engineering, University of Washington, Seattle, WA \\
$^\spadesuit$Allen Institute for Artificial Intelligence, Seattle, WA\\
$^\clubsuit$Microsoft Research, Redmond, WA \\
\texttt{\{antoineb, yejin\}@cs.washington.edu} \: \: \: \texttt{\{asli\}@ieee.org}}
\date{}

\begin{document}
\maketitle
\begin{abstract}

Large-scale learning of transformer language models has yielded improvements on a variety of natural language understanding tasks.
Whether they can be effectively adapted for summarization, however, has been less explored, as the learned representations are less seamlessly integrated into existing neural text production architectures. 
In this work, we propose two solutions for efficiently adapting pretrained transformer language models as text summarizers: source embeddings and domain-adaptive training.  
We test these solutions on three abstractive summarization datasets, achieving new state of the art performance on two of them. Finally, we show that these improvements are achieved by producing more focused summaries with fewer superfluous  and that performance improvements are more pronounced on more abstractive datasets.

\end{abstract}

\section{Introduction}
\label{sec:intro}
Recent work in large-scale language models \citep{elmo,gpt,bert} has allowed pretrained contextual representations to be easily adapted for a variety of downstream tasks, yielding improvements on many benchmarks evaluating natural language understanding \citep{glue}. Less explored, however, has been the effect of these pretrained representations on text production tasks, such as abstractive summarization, where state of the art performance is still achieved with sequence to sequence (seq2seq) models \citep{dca,gehrmann2018bottom}.

These sequence-to-sequence methods typically use an encoder and decoder model with separate parameters to represent the input article and produce the output summary, and the most successful solutions \citep{dca,gehrmann2018bottom,pointer-generator} use attention mechanisms that learn an alignment between encoder and decoder states. Pretrained language models, however, do not learn the parameters for such a task-specific alignment, making it challenging to integrate their learned representations into a summarization architecture at a higher level of abstraction than the word embedding. 

In this work, we adapt full transformer language models for abstractive summarization. Building off the work of \citet{Liu2018GeneratingWB}, who first proposed concatenating input and output text to a joint sequence and using a common transformer to encode both, we use a language model as a summarizer (rather than an encoder-decoder). With this approach, representations from a pretrained transformer language model (in this case, GPT \citep{gpt}) can be used to fully initialize the parameters of the summarization model, allowing it to leverage the representational power of a model trained at much larger scale. 

To accomplish this effectively, we outline two strategies for adapting pretrained representations for abstractive summarization. In the first, we augment the input representation of the summarization model by instantiating source embeddings that encode the token type of the text being read. 
This change allows the model to recognize whether a given token belongs to the input article or the output summary, thereby learning how to distinguish both types of text when encoding. 
In the second, we introduce a domain-adaptive training procedure that fine-tunes the transformer toward understanding general newswire text before training on the summarization end task directly, 
allowing the model to learn the general structure and language distribution of newswire text before being fine-tuned to produce summaries.

A comprehensive empirical study across three datasets, CNN/DailyMail \citep{cnndm}, XSum \citep{xsum}, and Newsroom \citep{newsroom}, shows that transformer language models can be used to train abstractive summarizers, producing summaries that are more concise and focused than state of the art baselines. Our investigation also empirically validates several observations about the abstractive summarization task. 
First, echoing the results of \citet{suncompare}, the most common summarization evaluation metric, ROUGE \citep{rougeA}, is highly sensitive to summary length, providing an advantage to methods that produce longer summaries, either through learning or with minimum summary length constraints. 
Second, achieving higher ROUGE scores is not strongly consistent with human assessments of abstractive summary quality.
Finally, despite being conceived as abstractive summarizers, most current state of the art models are highly extractive, copying phrases and even sentences verbatim from the document. 

\begin{figure*}[t]
    \centering
    \includegraphics[trim=7cm 10cm 6cm 6cm, width=0.25\columnwidth]{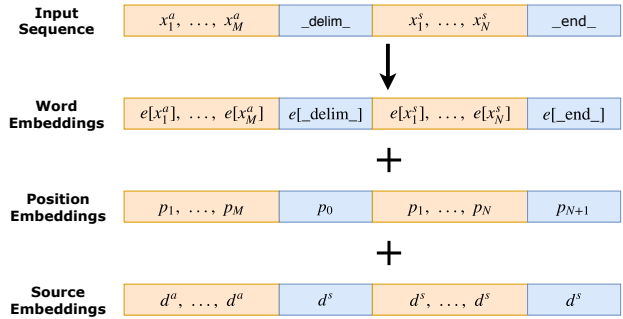}
    \caption{The embedding process for inputs to the \modelname~model.}
    \label{fig:transformer-summarization-input}
\end{figure*}

\section{Model}
\label{sec:model}
In this paper, we focus on a variant of the Transformer \citep{original-transformer} that has been pretrained on a large corpus of natural language stories: the GPT model \citep{gpt}. As our architecture is practically identical to the one proposed in \citet{gpt}, we point readers to that work for background on the architecture of the model, and focus below on the enhancements to the input representation made in our approach.

\subsection{Input Representation}
\label{ssec:model:input}

Each article is represented as a sequence of $M$ tokens $\mathbf{X}^a = \{x^a\}_{m=1}^M = x^a_{1}, ..., x^a_{M}$ and its corresponding summary is a sequence of $N$ tokens $\mathbf{X}^s = \{x^s\}_{n=1}^N = x^s_{1}, ..., x^s_{N}$. As outlined in Figure~\ref{fig:transformer-summarization-input}, the input structure of the training set is a pair of article and corresponding summary concatenated into two sequences similar to \citep{Liu2018GeneratingWB}:

\begin{equation}
    \mathbf{X} = \{x\}_{t=1}^T = [\mathbf{X}^a ; {<}\text{D}{>}; \mathbf{X}^s, {<}\text{E}{>}]
\end{equation} 
where $T$ = $M+N+2$, and $<$D$>$ and $<$E$>$ are special tokens identifying the delimitation and end of the sequence. Below, we define the process of encoding these sequences as inputs to the transformer.

\paragraph{Word Embedding} First, each token $x_t$ in the concatenated sequence $\mathbf{X}$ indexes a word embedding $e[x_t] \in \mathbb{R}^h $ 
from a joint vocabulary for the article and summary (and special tokens).

\paragraph{Position Embedding} Second, since the transformer (a self-attention model) has no concept of ordering of tokens, a position embedding $p_t \in \mathbb{R}^h$ is initialized for each absolute position in the sequence \citep{original-transformer}. The embedding for each position in the sequence is added to the word embedding of the token occupying that position, augmenting the final representation of the input. For example, each token in the article would be represented as: $w^a_m = e[x_m^a] + p_m$. Once the delimitation token $<$D$>$ is reached, the position counter is reset. For example, the first token of the article, $x^a_1$, and the first token of the summary, $x^s_1$, both receive $p_1$ as a positional embedding to augment their representations.

\paragraph{Source Embedding}
Finally, because the transformer must recognize pragmatic differences between the text of the article it reads and the text of the summary it learns to produce, an additional, source-specific embedding is initialized, $d \in \mathbb{R}^h$. 
The source embedding encodes whether a token is from the article portion $d^a$ of the concatenated input, or the summary portion $d^s$. For any article token (Eq.~\ref{eq:article_input}) or summary token (Eq.~\ref{eq:summary_input}) then, the final encoding is:

\begin{minipage}{0.45\linewidth}  
\begin{equation}
    \underset{m \in [1, M]}{\forall} w^a_m = e[x_m^a] + p_m + d^a
    \label{eq:article_input}
\end{equation}
\end{minipage}  
\hspace{0.5cm}  
\begin{minipage}{0.45\linewidth}  
\begin{equation}
    \underset{n \in [1, N]}{\forall} w^s_n = e[x_n^s] + p_n + d^s
    \label{eq:summary_input}
\end{equation}
\end{minipage}

\noindent In contrast to the other embeddings in the model, the source embeddings are not pretrained, introducing the potential that they could dominate pretrained representations for the word and position embeddings when summed (Eq.~\ref{eq:article_input},~\ref{eq:summary_input}). To avoid this, we normalize the random initialization of the source embeddings to have norm equal to half of the average norm of the word embeddings.
\section{Training}
The model is initialized with pretrained parameters from the GPT model \citep{gpt} that was trained on the BooksCorpus \citep{tbooks}. Following this initialization, we pursue two additional training procedures: domain-adaptive training and end task training.

\subsection{Domain-adapative Training}
Despite the benefit of using pretrained representations from the GPT model to initialize a summarizer, there is a language shift between the storybooks data on which the GPT model was trained and the type of language found in newswire summarization datasets \citep{cnndm,xsum,newsroom}. Additionally, there are structural differences between how articles are written (usually expressing salient points early on, followed by details later) and how stories unfold (less front-loading of key information).

To address this discrepancy, we propose domain-adaptive training (DAT) to adapt the transformer summarization model to the language distribution of newswire text by maximizing the conditional loglikelihood of the article tokens and summary tokens given all previous tokens in their concatenated input representation (see Figure~\ref{fig:transformer-summarization-input}):

\begin{equation}
    \mathcal{L}_{dat} = - \sum_{m=1}^M \log P(x_m^a \vert \{x^a\}_{1}^m) - \sum_{n=1}^N \log P(x_n^s \vert \{x^s\}_{1}^n, \{x^a\}_1^M) \label{eq:dat}
\end{equation}

\noindent where $M$ is length of the article, $N$ is the length of the summary, $\{x^a\}_{<m}$ is the set of all tokens in the article that precede $x^a_m$, $\{x^s\}_{<n}$ is the set of all tokens in the summary that precede $x^s_n$, and $\{x^a\}_M$ is the set of all article tokens. In this framework, the model is adapted to produce newswire-like language before being trained on the summarization end task, which only focuses on learning for summary production.

\subsection{End Task Training}
During end task training (ETT), the model is trained specifically to be able to produce a summary given a document, constraining the loss function toward maximizing the conditional loglikelihood of producing only the correct summary tokens given the set of article tokens $\{x^a\}_M$: 

\begin{equation}
    \mathcal{L}_{ett} = - \sum_{n=1}^N \log P(x_n^s \vert \{x^s\}_{<n}, \{x^a\}_M) \label{eq:ft}
\end{equation}

\noindent where $\{x^s\}_{<n}$ is the set of tokens in the summary that precede $x^s_n$.
\section{Experimental Setup}
\label{sec:setup}


\paragraph{Datasets}
The CNN/Daily Mail dataset \citep{cnndm} consists of articles from CNN 
and Daily Mail. 
Each article is associated with several descriptive bullet point highlights. Similar to previous work \citep{Nallapati2016AbstractiveTS}, we concatenate the highlights to create a target summary for each article in the dataset and use the same dataset splits. 
The Extreme Summarization (XSum) dataset \citep{xsum} consists of $\sim$230k article summary pairs taken from the BBC. 
 Each summary is a single sentence long and is professionally written (usually by the author), making the dataset exhibit more abstractive content than typical summarization datasets, such as CNN/DailyMail \citep{cnndm}. The Newsroom dataset \citep{newsroom} consists of $\sim$1.2M article summary pairs scraped from the Internet Archive. 
The articles 
 come from a set of 38 publishers and cover diverse topics. 
We provide statistics about each dataset in Table~\ref{tab:dataset-statistics}.

\paragraph{Data Preprocessing}
We used a bytepair encoding (BPE) for tokenization. For each summarization dataset, we use the BPE to tokenize each article and summary, and then truncate the articles to a maximum length of 512 tokens and each summary to a maximum length of 110 tokens. We then format each article summary pair into the format outlined in Figure \ref{fig:transformer-summarization-input}. 

\begin{table*}[t!]
    \centering
    \caption{Comparison of summarization datasets with respect to dataset size, proportion of unique n-grams, mean article length in words, and mean summary length in words.}
    \vspace{5pt}
    \resizebox{\columnwidth}{!}{
        \begin{tabular}{l|rrr|rrrr|rr}
            \toprule
            \multirow{2}{*}{Dataset}
             & \multicolumn{3}{c}{Split Size} & \multicolumn{4}{c}{\% Novel n-grams in Gold Summary} & \multicolumn{2}{c}{Mean \# Words} \\
             & Train & Validation & Test & unigrams & bigrams & trigrams & 4-grams & Article & Summary \\
            \toprule
            Newsroom & 993,746 & 108,590 & 108,636 & 17.40 & 44.05 & 55.38 & 61.21 & 658.6 & 26.7 \\
            XSum & 204,045 & 11,332 & 11,334 & 34.88 & 78.78 & 92.03 & 96.80 & 431.1 & 23.3 \\
            CNN/DailyMail & 287,227 & 13,368 & 11,490 & 12.70 & 46.29 & 65.04 & 75.56 & 685.2 & 52.0 \\
            \bottomrule
        \end{tabular}
    }
    \label{tab:dataset-statistics}
\end{table*}

\vspace{-1mm}
\paragraph{Model Specifications} We used a transformer decoder with $N=12$ blocks and $h=12$ masked self-attention heads in each block. We set the dimensionality of each self-attention head to be $d_{model}=768$. Unless stated otherwise, we use the pretrained weights of \citet{gpt} to initialize the parameters of the model. 
Special tokens that are added to the vocabulary (i.e. the end token, start token, and delimiter token) are initialized by sampling from the standard normal distribution. Our full model with source embeddings (\S\ref{ssec:model:input}) is denoted as as \modelname~and we also train an ablation, Transformer-LM, that does not use source embeddings.
\vspace{-1mm}
\paragraph{Training Details}
All models were trained with a learning rate of $6.25 \times 10^{-5}$ and a minibatch size of 64.  
When domain-adaptive training (DAT) is used, we train for 10 epochs using DAT and then for an additional 10 epochs using end task training (ETT). Without DAT, we train on the end task for 20 epochs. Unless specified otherwise, the final model trained for each dataset uses both domain-adaptive training and end task training. We did not tune hyperparameters. All models were trained using the PyTorch package\footnote{\url{https://pytorch.org/}}
and the HuggingFace implementation of GPT.\footnote{\url{https://github.com/huggingface/pytorch-openai-transformer-lm}} We trained each model on 8 Tesla V100-SMX2. Training for a total of 20 epochs took approximately 1 day of clock time for the XSum and CNN/Daily Mail datasets, and 3 days for the Newsroom dataset. Our source code is publicly available.\footnote{\url{https://github.com/Andrew03/transformer-abstractive-summarization}}

\vspace{-1mm}
\paragraph{Generation} We perform generation by using beam search with a beam size of 3. We use the trigram trick \citep{rlsummpaulus} during beam search. Each summary token is generated by decoding from the distribution yielded by the model from processing an input tensor that is the concatenation of the article tokens, the delimiter token, and any previously generated summary tokens.
\vspace{-1mm}
\paragraph{Evaluation}
We evaluate our system with common summarization metrics: ROUGE-1 (R-1), a measure of unigram recall between the summary and document, ROUGE-2 (R-2), a similar measure of bigram recall, and ROUGE-L (R-L), a measure of the longest common subsequence between the summary and document \cite{lin2004rouge}. We also report the length of the summary in terms of tokens produced. For each dataset, for evaluation on the test set, we selected models with the largest ROUGE-1 score on a subset of 500 samples from the validation set.
\section{Experiments}

\subsection{CNN/Daily Mail}
\paragraph{Baselines}
We report the results from various models previously trained and evaluated on the CNN/Daily Mail dataset. The PGen and PGen + Coverage models \citep{pointer-generator}, consist of attentive RNN encoder-decoders that integrate the ability to directly copy from the article when generating tokens. \citet{pasunuru2018multi} extend this work by adding policy gradient training with a mixture of rewards that promote saliency and entailment. Bottom-up summarization and the Copy Transformer \citep{gehrmann2018bottom} also extend \citet{pointer-generator} by using the copy mechanism to compress the article to only relevant content before summarizing it. \citet{chen2018fast} also look at performing content selection, but extract full sentences from the document with a novel extractor model. 
Finally, the DCA model \citep{dca} uses multiple separate communicating encoders over different parts of the document to produce representations that are more focused on salient details.

\begin{table*}[h]
    \centering
    \caption{\textbf{ROUGE F1} results on the test set of CNN/Daily Mail. 
    Best model results are bolded.
}
    \vspace{5pt}
        \begin{tabular}{l rrrr}
            \toprule
            Model & R-1 & R-2 & R-L & Length ($L$) \\
            \toprule
            PGen \citep{pointer-generator} & 36.44 & 15.66 & 33.42 & 53.69\\
            PGen + Coverage \citep{pointer-generator} & 39.53 & 17.28 & 36.38 & 59.75\\
            RougeSal + Ent RL \citep{pasunuru2018multi} & 40.43 & 18.00 & 37.10 & - \\
            Bottom-Up Summ \citep{gehrmann2018bottom} & 41.22 & 18.68 & \textbf{38.34} & 55.25 \\
            CopyTransformer \citep{gehrmann2018bottom}& 40.96 & 18.38 & 38.16 & -  \\
            rnn-ext + RL \citep{chen2018fast} & 41.47 & 18.72 & 37.76 & 77.44 \\
            DCA \citep{dca} & \textbf{41.67} & \textbf{19.47} & 37.92 & 51.01\\
            \midrule
            Transformer-LM & 38.67 & 17.47 & 35.79 & 43.40 \\
            \modelname & 37.96 & 17.36 & 35.12 & 42.42 \\
            \bottomrule
        \end{tabular}
    \label{tab:cnndm-results}
\end{table*}

\paragraph{Automatic Metrics}
We report our results using automatic metrics in Table \ref{tab:cnndm-results}.
On this dataset, our main model, \modelname, performs slightly worse than other state of the art models. 
We note that our model tends to generate shorter summaries than the gold summaries ($\sim 20\%$ shorter), which could lower ROUGE recall performance. 

\begin{figure*}[h]
    \centering
    \includegraphics[width=\columnwidth, height=5.5cm]{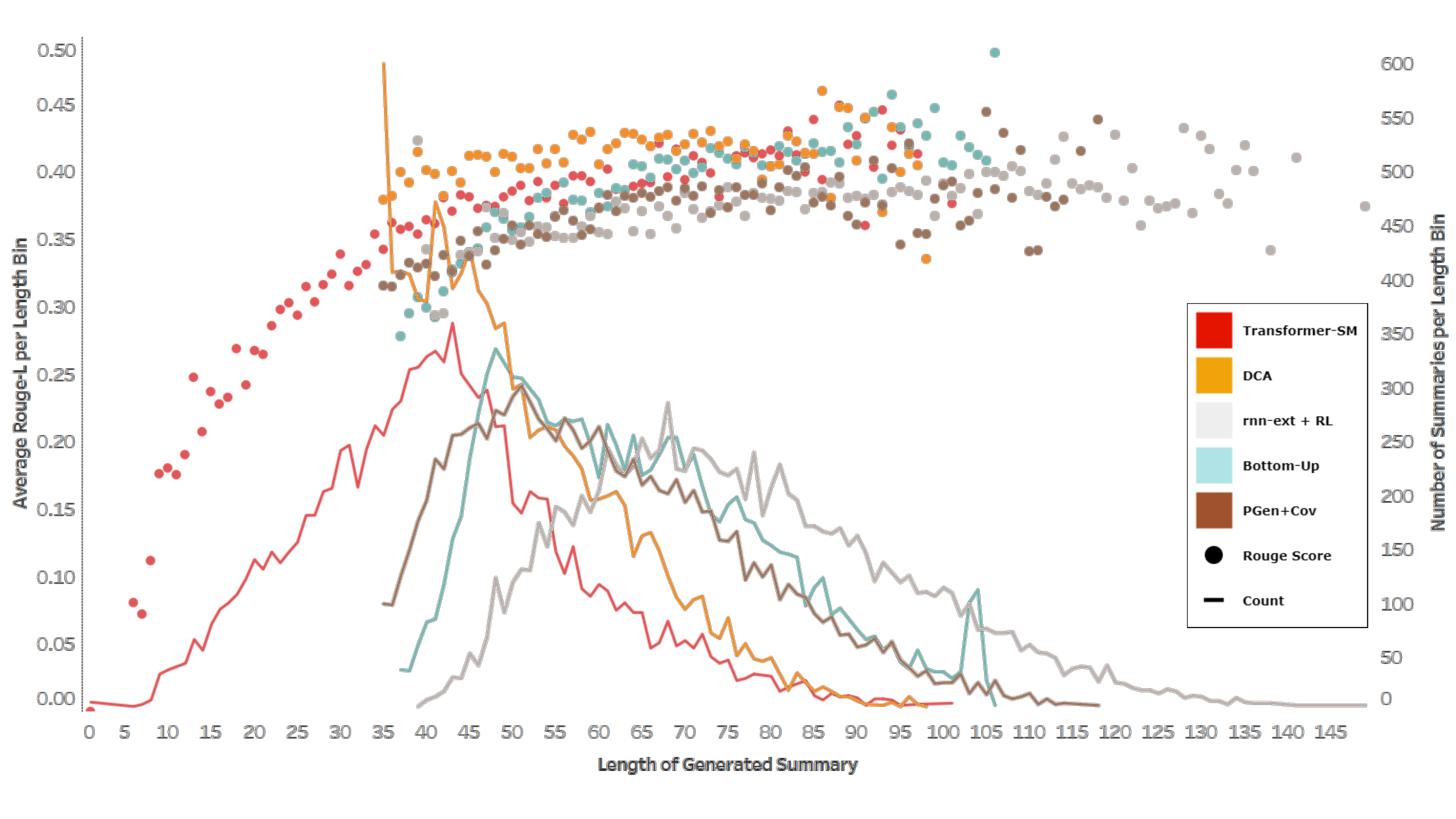}
    \caption{Average ROUGE-L for summaries in different length bins. Scatter plots correspond to ROUGE-L scores for each bin, while solid lines correspond to the number of summaries in each bin
 }
    \label{fig:length_v_rouge}
\end{figure*}

In Figure~\ref{fig:length_v_rouge},
we investigate the correlation of ROUGE-L scores with summary length, and note that a minimum decoding length used by 
state-of-the-art algorithms places baseline generated summaries in length bins of higher average ROUGE-L performance. When \modelname~produces summaries in these same length bins (i.e., more than 30 tokens), its performance is only consistently beaten by the DCA model, which was fine-tuned with RL.

\vspace{-5pt}
\paragraph{Human Evaluation}
While ROUGE scores are negatively influenced by the shorter average length of the summaries produced by our model, it is not clear that shorter summaries are correlated with worse quality. To evaluate this hypothesis, we perform a human evaluation on 125 (article, summary) pairs randomly sampled from the test set. The article and model-generated summaries were presented to three workers from Amazon Mechanical Turk (AMT). 

\begin{table}[t]
    \centering
    \caption{Head-to-head comparison between test set outputs of (Left) DCA \citep{dca} and \modelname ~(Right) PGen+Cov \citep{pointer-generator} and \modelname. Analyses done on summaries for CNN/DailyMail.}
        \begin{tabular}{l rrr | rrr}
            \toprule
            Model           & DCA   & Same  & \modelnameshort& PGen+Cov & Same  & \modelnameshort \\
            \toprule
            Non-redundancy  & 96    & 116   & \textbf{163}   & 77       & 85    & \textbf{213}    \\
            Coherence       & 136   & 60    & \textbf{179}   & 160      & 35    & \textbf{180}    \\
            Focus           & 136   & 36    & \textbf{203}   & 115      & 33    & \textbf{227}    \\
            \midrule
            Overall         & 138   & 36    & \textbf{201}   & 150      & 39    & \textbf{186}    \\
            \bottomrule \\
        \end{tabular}
    \label{tab:cnndm-human-results}
\end{table}

Each worker was presented two model-generated summaries, one produced by the \modelname~model, and one from the DCA model \citep{dca} or the PGen+Cov model \citep{pointer-generator}. Workers were asked to select the better summary for four different quality metrics from \citet{dca}: \textit{non-redundancy} (fewer of the same ideas are repeated), \textit{coherence} (ideas are expressed clearly), \textit{focus} (the main ideas of the document are shared while avoiding superfluous details), and \textit{overall}. 

The results are presented in Table \ref{tab:cnndm-human-results}. Interestingly, the summaries from \modelname~are consistently preferred by humans across all 4 evaluations dimensions compared to those from the DCA and PGen+Coverage models, indicating that the \modelname's lower ROUGE scores observed in Table~\ref{tab:cnndm-results} are not necessarily correlated with human judgments of quality. 

\begin{table}[h]
\begin{minipage}{0.47\linewidth}
    \centering
    \caption{ROUGE-L precision (R-L P), recall (R-L R), and F1 (R-L F1) scores computed between generated summaries and input CNN/DailyMail articles after removing stop words}
    \vspace{5pt}
         \begin{tabular}{l rrr}
            \toprule
            Model Name
            & R-L P & R-L R &  \multicolumn{1}{c}{$L$} \\  
            \toprule
            PGen \citep{pointer-generator}      & 95.22 & 8.22  & 53.69 \\ 
            PGen+Cov \citep{pointer-generator}  & 99.83 & 10.74 & 59.75 \\ 
            Bottom-Up \citep{gehrmann2018bottom}& 98.93 & 9.35  & 55.25 \\ 
            rnn-ext + RL \citep{chen2018fast}   & 99.05 & 12.77 & 77.44 \\ 
            DCA \citep{dca}                     & 97.31 & 8.24  & 51.01 \\ 
            Transformer-LM                      & 96.66 & 9.95  & 43.40 \\ 
            \modelname                          & 97.16 & 9.78  & 42.42 \\ 
            \midrule
            Gold Summary                        & 79.88 & 11.13 & 52.02\\ 
            \bottomrule \\
        \end{tabular}
    \label{tab:cnndm-efficiency}
\end{minipage}
\hspace{0.3cm}
\begin{minipage}{0.48\linewidth}
    \centering
    \caption{Ablation study of training schedules on CNN/DailyMail. (PT) Model initialized with pretrained weights; (DAT) Model uses domain-adaptive training; (ETT) trained on end task.}
    \vspace{5pt}
        \begin{tabular}{l rrr}
            \toprule
            Model & R-1 & R-2 & R-L  \\
            \toprule
            T-LM ({\footnotesize ETT}) & 36.82 & 16.04 & 34.03 \\
            T-LM ({\footnotesize DAT+ETT}) & 38.00 & 17.13 & 35.20 \\
            T-LM ({\footnotesize PT+ETT}) & 38.20 & 17.39 & 35.40 \\
            T-LM ({\footnotesize PT+DAT+ETT}) & \textbf{39.01} & \textbf{17.87} & \textbf{36.17} \\
            \midrule
            \modelnameshort~({\footnotesize ETT}) & 37.81 & 16.82 & 34.87 \\
            \modelnameshort~({\footnotesize DAT+ETT}) & 38.34 & 17.34 & 35.44 \\
            \modelnameshort~({\footnotesize PT+ETT}) & \textbf{38.71} & 17.53 & \textbf{35.90} \\
            \modelnameshort~({\footnotesize PT+DAT+ETT}) & 38.33 & \textbf{17.79} & 35.56 \\
            \bottomrule \\
        \end{tabular}
    \label{tab:adaptive-vs-fine-tuning}
\end{minipage}
\end{table}
\vspace{-5pt}
\paragraph{Efficiency}
Due to the large improvements over the baseline models in the human evaluation categories of \emph{non-redundancy} and \emph{focus}, and the generally shorter summaries produced by \modelname, 
we investigate whether \modelname~is able to more efficiently express key ideas of the document. To evaluate the efficiency of each model, we remove non-content words from the model-generated summaries and articles, and compute the ROUGE score between them. This measure serves as a proxy for the rate at which ideas expressed in the summary can be found in the document.

We report these results in Table~\ref{tab:cnndm-efficiency} and observe that \modelname~reports comparable ROUGE-L recall scores to other baselines when evaluated with respect to the article, despite producing summaries that, on average, 27\% shorter. Meanwhile, ROUGE-L precision is also very similar to the baseline models, indicating that the summaries of all models indicate a similar degree of information relevance.\footnote{The high precision scores across all models confirm that despite being conceived as abstractive generators, these models display highly extractive behavior.} 
Combined with the results from Table~\ref{tab:cnndm-human-results}, we conjecture that \modelname~is able to more efficiently express key ideas from the document. While other models may be producing longer summaries that yield higher ROUGE performance (Table~\ref{tab:cnndm-results}), the additional tokens may reflect redundant and unsalient information, which human evaluators penalize. 

\paragraph{Analysis of domain-adaptive training and source embeddings} Our approach involved two strategies for efficiently using transformer language models for abstractive summarization: domain-adaptive training and source embeddings. To assess their individual impact, we evaluate multiple training schedule permutations (e.g., various combinations of using pretrained representations from the GPT model and using domain-adaptive training), as well as the impact of source embeddings. Our results in Table~\ref{tab:adaptive-vs-fine-tuning} yield multiple interesting conclusions. First, in general, domain-adaptive training (+DAT in Table~\ref{tab:adaptive-vs-fine-tuning}) provides a clear improvement over training directly on the end task, irrespective of whether pretrained representations are used. Similarly, using source embeddings (T-SM in Table~\ref{tab:adaptive-vs-fine-tuning}) provides a repeated improvement over the T-LM ablation. Surprisingly, when pretrained initializations, DAT, and source embeddings are used in tandem, performance drops slightly compared to not using DAT or not using source embeddings. We note, however, that this observation does not hold true for the XSum dataset (\S\ref{ssec:experiments:xsum}), and conjecture that the extractive nature of the CNN/DailyMail dataset may make these approaches have redundant effects in this setting.

\subsection{XSum}
\label{ssec:experiments:xsum}
A study on the quality of abstractive summaries is best performed on the XSum dataset \citep{xsum}, which is specifically designed with gold summaries that are less extractive than the other datasets (Table~\ref{tab:dataset-statistics}).

\paragraph{Baselines}
We report the performance of \modelname~on this dataset in comparison to baselines originally reported in \citet{xsum}: an attention-based sequence to sequence model (AttnS2S), a pointer-generator model capable of generating words and copying directly from the input (PGen), a second pointer-generator model with a coverage mechanism to prevent repetition (PGen+Cov), and the top performing variant of the topic aware convolutional sequence to sequence model (T-ConvS2S), in which the encoder and decoder are provided with word topic and document topic distributions obtained using LDA as additional inputs. Our final baseline is the Multi-Level Memory Network (MMN) \citep{kim2018abstractive}, which applies attention over multiple memory layers for varying levels of abstraction.
\begin{wraptable}{R}{0.46\textwidth}
\vspace{-15pt}
    \centering
    \caption{Comparison results on the XSum test set using the \textbf{F1} variants of \textbf{ROUGE}}
    \begin{tabular}{l rrr}
        \toprule
        Model & R-1 & R-2 & R-L  \\
        \toprule
        AttnS2S \citep{xsum} & 28.42 & 8.77 & 22.48 \\
        PGen \citep{xsum} & 29.70 & 9.21 & 23.24 \\
        PGen+Cov \citep{xsum} & 28.10 & 8.02 & 21.72 \\
        T-ConvS2S \citep{xsum} & 31.89 & 11.54 & 25.75 \\
        MMN \citep{kim2018abstractive} & 32.00 & 12.10 & 26.00 \\
        \midrule
        Transformer-LM & 36.03 & 14.57 & 29.20 \\
        \modelname & \textbf{36.73} & \textbf{14.93} & \textbf{29.66} \\
        \bottomrule
    \end{tabular}
    \label{tab:xsum-results}
\vspace{-5pt}
\end{wraptable}
\paragraph{Results}
We report our results in Table \ref{tab:xsum-results}. Our models significantly outperform the comparison baselines across all three variants of the ROUGE metric. Interestingly, the \modelname~ achieves noticeable improvement over the Transformer-LM model, suggesting that both source embeddings and domain adaptive training are helpful when target summaries are more abstractive. Examples of model-generated summaries from the XSum dataset illustrate the improvement over baselines qualitatively in Table~\ref{tab:examples}. 
In support of results presented earlier, the model produces abstractive summaries that provide focused information about the main points of the articles.

\begin{table*}[t]
    \centering
    \small
    \caption{XSum samples for the baseline T-ConvS2S model and \modelname~along with the gold summary. Articles are shortened for brevity. Capitalization was manually added for ease of reading.
}
\vspace{10pt}
\resizebox{\columnwidth}{!}{
    \begin{tabular}{lL}
        \toprule
        Source & Source Text \\
        \toprule
        Article snippet & \small{Officials said the attack happened at the Europa shopping centre in the capital Minsk. ... 
        Police later arrested the 18-year-old suspect. 
        ... "He cut one woman with the chainsaw and hit her with a hammer. She died. He also attacked others." The injured woman was taken to a local hospital. The attacker had brought the chainsaw and the axe to the shopping centre ...} \\
        \midrule
        T-ConvS2S & A man has been arrested on suspicion of attempted murder by after a knife attack on a shopping centre in central London. \\
        \modelname & A teenage girl has been killed by a chainsaw attack at a shopping centre in central Russia, police say. \\
        \midrule
        Gold & A young man has attacked people with a chainsaw and an axe at a shopping centre in Belarus, killing one woman and injuring another. \\
        \toprule
        \toprule
        Article snippet & \small{The 34-year-old Sweden striker's contract with the french champions expires in the summer, and he has been linked with Manchester United, LA Galaxy and AC Milan. ... 
        PSG said Ibrahimovic leaves as "the greatest striker and one of the very best players in the club's history" . ...
        } \\
        \midrule
        T-ConvS2S & Paris St-Germain have completed the signing of Zlatan Ibrahimovic from Paris St-Germain for an undisclosed fee. \\
        \modelname & Zlatan Ibrahimovic says he will leave Paris St-Germain at the end of the season to return to the club. \\
        \midrule
        Gold & Zlatan Ibrahimovic will leave Paris St-Germain at the end of the season. \\
        \toprule
        \toprule
        Article snippet & \small{
        ... The animal was taken from Lathom pets and aquatics in Ormskirk on Tuesday afternoon, Lancashire police said. 
        The shop's owner said CCTV showed a man taking the tortoise - which needs calcium supplements - out of the tank. ...
        } \\
        \midrule
        T-ConvS2S & A puppy's pet shop has been stolen from a shop in Lancashire. \\
        \modelname & A tortoise has been stolen from a pet shop. \\ 
        \midrule
        Gold & A baby tortoise has been stolen from a pet shop in Lancashire. \\
        \bottomrule
    \end{tabular}
    }
    \label{tab:examples}
\end{table*}

\subsection{Newsroom}
Finally, we report the performance of our model on the Newsroom dataset \citep{newsroom}, the largest of the evaluation datasets. Due to the large cost of training, only the \modelname~model was evaluated.

\paragraph{Baselines}
As baselines, we report the performance of models released by the authors of the Newsroom dataset \citep{newsroom}. These models included an attentive encoder-decoder (Attn-S2S) and a pointer-generator network (PGen). We also compared against C10110 \citep{shi2018neural}, a complex encoder-decoder that uses LSTMs, encoder attention, intra-decoder attention, and pointer-generation to produce summaries.  We also compare against the Multi-Level Memory Network (MMN) \citep{kim2018abstractive} mentioned earlier. The authors of this baseline only evaluated on the abstractive subset of the Newsroom dataset.

\begin{wraptable}{L}{0.44\textwidth}
\vspace{5pt}
    \centering
    \caption{Comparison results on the Newsroom test set using \textbf{ROUGE} \textbf{F1} }
    \vspace{10pt}
        \begin{tabular}{l rrr}
            \toprule
            Model & R-1 & R-2 & R-L  \\
            \toprule
            Attn-S2S \citep{xsum} & 5.88 & 0.39 & 5.32 \\
            PGen \citep{xsum} & 26.02 & 13.25 & 22.43 \\
            C10110 \citep{shi2018neural} & 39.36 & 27.86 & 36.35 \\
            \midrule
            \modelnameshort & \textbf{40.87} & \textbf{28.59} & \textbf{37.62} \\
            \bottomrule
        \end{tabular}
    \label{tab:newsroom-results}
\end{wraptable}

\paragraph{Results}
We report our results with ROUGE-style automatic metrics in Table \ref{tab:newsroom-results}, showing that \modelname~outperforms the previous best model, C10110 \citep{shi2018neural}, across all metrics. Interestingly, our model achieves its highest performance increase over baseline models on Rouge-L, the metric usually considered as being most strongly correlated with strong summaries. 
Furthermore, an analysis of different validation subsets of the Newsroom dataset in Table~\ref{tab:newsroom-val-results} (split on the level of extractiveness of the gold summaries) shows that \modelname~performs better than baselines approaches on all varieties of summary types.

\begin{table*}
    \centering
    \caption{\textbf{ROUGE} \textbf{F1} results on validation subsets and full validation set for Newsroom}
    \vspace{5pt}
    \resizebox{\columnwidth}{!}{
        \begin{tabular}{lrrr|rrr|rrr|rrr}
            \toprule
            \multirow{2}{*}{Model Name}
            & \multicolumn{3}{c}{Extractive} & \multicolumn{3}{c}{Mixed} & \multicolumn{3}{c}{Abstractive} & \multicolumn{3}{c}{Newsroom-D} \\
            \cmidrule{2-4} \cmidrule{5-7} \cmidrule{8-10} \cmidrule{11-13}
            & R-1 & R-2 & R-L & R-1 & R-2 & R-L & R-1 & R-2 & R-L & R-1 & R-2 & R-L\\
            \toprule
            PGen \citep{newsroom} & 39.11 & 27.95 & 36.17 & 25.48 & 11.04 & 21.06 & 14.66 & 2.26 & 11.44 & 26.27 & 13.55 & 22.72 \\
            MMN \citep{kim2018abstractive} & - & - & - & - & - & - & 17.5 & 4.7 & 14.2 & - & - & - \\
            \modelname & \textbf{64.69} & \textbf{59.72} & \textbf{63.56} & \textbf{35.75} & \textbf{18.83} & \textbf{30.63} & \textbf{22.44} & \textbf{7.39} & \textbf{18.75} & \textbf{41.42} & \textbf{29.51} & \textbf{38.26} \\
            \bottomrule
        \end{tabular}
    }
    \label{tab:newsroom-val-results}
\end{table*}

\section{Related Work}
\label{sec:related}
\paragraph{Abstractive Summarization}
There has been a large variety of work exploring different methods for neural abstractive document summarization. Attention mechanisms have been shown to improve a variety of models \citep{recurrent-extractive-summarization, graph-summarization, neural-extractive-summarization}, and is one of the motivating factors for this work. Pointer generator networks introduced in \citet{pointer-generator} have been shown to increase summary veracity, and inspired the tangential usage of copy mechanisms in Transformers for document summarization \citet{gehrmann2018bottom}. Other works have also explored the use of reinforcement learning to directly optimize summarization models on the ROUGE metric \citep{pasunuru2018multi,rlsummpaulus,chen2018fast}.

\paragraph{Contextualized Representations}
Our approach is also relevant to recent work on contextualized language representations that are pretrained on large-scale language corpora. These representations can then be simply integrated or fine-tuned for improved performance on many downstream tasks \citep{glue}. SSL \cite{Dai2015SemisupervisedSL}, CoVe \cite{McCann2017LearnedIT}, and ELMo \citep{elmo} all learned contextualized representations through training RNN language models and encoder-decoders. Follow-up work extended these ideas, but replaced the RNN with a deep transformer \cite{gpt} that was trained to learn language patterns on a large story dataset. BERT \citep{bert} more clearly extended the idea of using Transformers for language modeling by making the encoded representations bidirectional and adding two new loss functions: a masked token loss and next sentence prediction loss for more accurate discourse representations. More recently, GPT2  \cite{gpt2} expanded the scale of pretrained language models, and showed promising results on zero-shot tasks.
\section{Conclusion}
\label{sec:conclusion}

In this work, we introduce two approaches for effectively adapting pretrained language model representations to abstractive summarization: domain-adaptive training, and source embeddings. We evaluate the effect of both approaches across three abstractive summarization testbeds: CNN/DailyMail, XSum, and Newsroom, and achieve state of the art ROUGE-L results on two of them, while showing superior human evaluation performance on the third. In the process, we show that the ROUGE-L metric often used for abstractive summarization evaluation is quite sensitive to summary length, allowing it to be exploitable by approaches that use heuristics to control summary length.

\bibliography{neurips2019}

\begin{thebibliography}{28}
\expandafter\ifx\csname natexlab\endcsname\relax\def\natexlab#1{#1}\fi

\bibitem[{Celikyilmaz et~al.(2018)Celikyilmaz, Bosselut, He, and Choi}]{dca}
Asli Celikyilmaz, Antoine Bosselut, Xiaodong He, and Yejin Choi. 2018.
\newblock Deep communicating agents for abstractive summarization.
\newblock In \emph{NAACL}.

\bibitem[{Chen and Bansal(2018)}]{chen2018fast}
Yen-Chun Chen and Mohit Bansal. 2018.
\newblock Fast abstractive summarization with reinforce-selected sentence
  rewriting.
\newblock In \emph{ACL}.

\bibitem[{Cheng and Lapata(2016)}]{neural-extractive-summarization}
Jianpeng Cheng and Mirella Lapata. 2016.
\newblock Neural summarization by extracting sentences and words.
\newblock \emph{arXiv preprint arXiv:1603.07252}.

\bibitem[{Dai and Le(2015)}]{Dai2015SemisupervisedSL}
Andrew~M. Dai and Quoc~V. Le. 2015.
\newblock Semi-supervised sequence learning.
\newblock In \emph{NIPS}.

\bibitem[{Devlin et~al.(2019)Devlin, Chang, Lee, and Toutanova}]{bert}
Jacob Devlin, Ming-Wei Chang, Kenton Lee, and Kristina Toutanova. 2019.
\newblock Bert: Pre-training of deep bidirectional transformers for language
  understanding.
\newblock In \emph{NAACL}.

\bibitem[{Gehrmann et~al.(2018)Gehrmann, Deng, and Rush}]{gehrmann2018bottom}
Sebastian Gehrmann, Yuntian Deng, and Alexander~M Rush. 2018.
\newblock Bottom-up abstractive summarization.
\newblock In \emph{EMNLP}.

\bibitem[{Grusky et~al.(2019)Grusky, Naaman, and Artzi}]{newsroom}
Max Grusky, Mor Naaman, and Yoav Artzi. 2019.
\newblock Newsroom: A dataset of 1.3 million summaries with diverse extractive
  strategies.
\newblock In \emph{NAACL}.

\bibitem[{Hermann et~al.(2015)Hermann, Kocisky, Grefenstette, Espeholt, Kay,
  Suleyman, and Blunsom}]{cnndm}
Karl~Moritz Hermann, Tomas Kocisky, Edward Grefenstette, Lasse Espeholt, Will
  Kay, Mustafa Suleyman, and Phil Blunsom. 2015.
\newblock Teaching machines to read and comprehend.
\newblock In \emph{Advances in Neural Information Processing Systems}, pages
  1693--1701.

\bibitem[{Kim et~al.(2018)Kim, Kim, and Kim}]{kim2018abstractive}
Byeongchang Kim, Hyunwoo Kim, and Gunhee Kim. 2018.
\newblock Abstractive summarization of reddit posts with multi-level memory
  networks.
\newblock \emph{arXiv preprint arXiv:1811.00783}.

\bibitem[{Lin(2004{\natexlab{a}})}]{rougeA}
Chin-Yew Lin. 2004{\natexlab{a}}.
\newblock Looking for a few good metrics: Automatic summarization
  evaluation-how many samples are enough?
\newblock In \emph{NTCIR}.

\bibitem[{Lin(2004{\natexlab{b}})}]{lin2004rouge}
Chin-Yew Lin. 2004{\natexlab{b}}.
\newblock Rouge: A package for automatic evaluation of summaries.
\newblock \emph{Text Summarization Branches Out}.

\bibitem[{Liu et~al.(2018)Liu, Saleh, Pot, Goodrich, Sepassi, Kaiser, and
  Shazeer}]{Liu2018GeneratingWB}
Peter~J. Liu, Mohammad Saleh, Etienne Pot, Ben Goodrich, Ryan Sepassi, Lukasz
  Kaiser, and Noam Shazeer. 2018.
\newblock Generating wikipedia by summarizing long sequences.
\newblock In \emph{ICLR}.

\bibitem[{McCann et~al.(2017)McCann, Bradbury, Xiong, and
  Socher}]{McCann2017LearnedIT}
Bryan McCann, James Bradbury, Caiming Xiong, and Richard Socher. 2017.
\newblock Learned in translation: Contextualized word vectors.
\newblock In \emph{NIPS}.

\bibitem[{Nallapati et~al.(2017)Nallapati, Zhai, and
  Zhou}]{recurrent-extractive-summarization}
Ramesh Nallapati, Feifei Zhai, and Bowen Zhou. 2017.
\newblock Summarunner: A recurrent neural network based sequence model for
  extractive summarization of documents.
\newblock In \emph{Thirty-First AAAI Conference on Artificial Intelligence}.

\bibitem[{Nallapati et~al.(2016)Nallapati, Zhou, dos Santos, Çaglar
  G{\"u}lçehre, and Xiang}]{Nallapati2016AbstractiveTS}
Ramesh Nallapati, Bowen Zhou, C{\'i}cero~Nogueira dos Santos, Çaglar
  G{\"u}lçehre, and Bing Xiang. 2016.
\newblock Abstractive text summarization using sequence-to-sequence rnns and
  beyond.
\newblock In \emph{CoNLL}.

\bibitem[{Narayan et~al.(2018)Narayan, Cohen, and Lapata}]{xsum}
Shashi Narayan, Shay~B Cohen, and Mirella Lapata. 2018.
\newblock Don't give me the details, just the summary! topic-aware
  convolutional neural networks for extreme summarization.
\newblock In \emph{EMNLP}.

\bibitem[{Pasunuru and Bansal(2018)}]{pasunuru2018multi}
Ramakanth Pasunuru and Mohit Bansal. 2018.
\newblock Multi-reward reinforced summarization with saliency and entailment.
\newblock In \emph{ACL}.

\bibitem[{Paulus et~al.(2018)Paulus, Xiong, and Socher}]{rlsummpaulus}
Romain Paulus, Caiming Xiong, and Richard Socher. 2018.
\newblock A deep reinforced model for abstractive summarization.
\newblock In \emph{ICLR}.

\bibitem[{Peters et~al.(2018)Peters, Neumann, Iyyer, Gardner, Clark, Lee, and
  Zettlemoyer}]{elmo}
Matthew~E. Peters, Mark Neumann, Mohit Iyyer, Matt Gardner, Christopher Clark,
  Kenton Lee, and Luke Zettlemoyer. 2018.
\newblock Deep contextualized word representations.
\newblock In \emph{Proc. of NAACL}.

\bibitem[{Radford et~al.(2018)Radford, Narasimhan, Salimans, and
  Sutskever}]{gpt}
Alec Radford, Karthik Narasimhan, Tim Salimans, and Ilya Sutskever. 2018.
\newblock Improving language understanding by generative pre-training. 2018.
\newblock \emph{URL https://s3-us-west-2. amazonaws.
  com/openai-assets/research-covers/language-unsupervised/language\_understanding\_paper.
  pdf}.

\bibitem[{Radford et~al.(2019)Radford, Wu, Child, Luan, Amodei, and
  Sutskever}]{gpt2}
Alec Radford, Jeff Wu, Rewon Child, David Luan, Dario Amodei, and Ilya
  Sutskever. 2019.
\newblock Language models are unsupervised multitask learners.

\bibitem[{See et~al.(2017)See, Liu, and Manning}]{pointer-generator}
Abigail See, Peter~J Liu, and Christopher~D Manning. 2017.
\newblock Get to the point: Summarization with pointer-generator networks.
\newblock In \emph{ACL}.

\bibitem[{Shi et~al.(2018)Shi, Keneshloo, Ramakrishnan, and
  Reddy}]{shi2018neural}
Tian Shi, Yaser Keneshloo, Naren Ramakrishnan, and Chandan~K Reddy. 2018.
\newblock Neural abstractive text summarization with sequence-to-sequence
  models.
\newblock \emph{arXiv preprint arXiv:1812.02303}.

\bibitem[{Sun et~al.(2019)Sun, Shapira, Dagan, and Nenkova}]{suncompare}
Simeng Sun, Ori Shapira, Ido Dagan, and Ani Nenkova. 2019.
\newblock How to compare summarizers without target length? pitfalls, solutions
  and re-examination of the neural summarization literature.
\newblock In \emph{Proceedings of NAACL 2019 Workshop on Optimizing and
  Evaluating Neural Language Generation (NeuralGen)}.

\bibitem[{Tan et~al.(2017)Tan, Wan, and Xiao}]{graph-summarization}
Jiwei Tan, Xiaojun Wan, and Jianguo Xiao. 2017.
\newblock Abstractive document summarization with a graph-based attentional
  neural model.
\newblock In \emph{Proceedings of the 55th Annual Meeting of the Association
  for Computational Linguistics (Volume 1: Long Papers)}, volume~1, pages
  1171--1181.

\bibitem[{Vaswani et~al.(2017)Vaswani, Shazeer, Parmar, Uszkoreit, Jones,
  Gomez, Kaiser, and Polosukhin}]{original-transformer}
Ashish Vaswani, Noam Shazeer, Niki Parmar, Jakob Uszkoreit, Llion Jones,
  Aidan~N Gomez, {\L}ukasz Kaiser, and Illia Polosukhin. 2017.
\newblock Attention is all you need.
\newblock In \emph{Advances in Neural Information Processing Systems}, pages
  5998--6008.

\bibitem[{Wang et~al.(2018)Wang, Singh, Michael, Hill, Levy, and Bowman}]{glue}
Alex Wang, Amanpreet Singh, Julian Michael, Felix Hill, Omer Levy, and
  Samuel~R. Bowman. 2018.
\newblock Glue: A multi-task benchmark and analysis platform for natural
  language understanding.
\newblock \emph{arXiv preprint 1804.07461}.

\bibitem[{Zhu et~al.(2015)Zhu, Kiros, Zemel, Salakhutdinov, Urtasun, Torralba,
  and Fidler}]{tbooks}
Yukun Zhu, Ryan Kiros, Richard~S. Zemel, Ruslan~R. Salakhutdinov, Raquel
  Urtasun, Antonio Torralba, and Sanja Fidler. 2015.
\newblock Aligning books and movies: Towards story-like visual explanations by
  watching movies and reading books.
\newblock \emph{2015 IEEE International Conference on Computer Vision (ICCV)},
  pages 19--27.

\end{thebibliography}
\bibliographystyle{acl_natbib}
\newpage
\appendix

\section{Reproducibility}

We provide additional details relevant to the experimental environment here.

\paragraph{Data Sources}
The CNN/Daily Mail dataset \citep{cnndm} consists of articles from CNN 
and Daily Mail.\footnote{\url{https://www.cnn.com};\hspace{5pt} \url{https://www.dailymail.co.uk}} 
Each article is associated with several descriptive bullet point highlights. Similar to previous work \citep{Nallapati2016AbstractiveTS}, we concatenate the highlights to create a target summary for each article in the dataset.
The Newsroom dataset \citep{newsroom} consists of $\sim$1.2M article summary pairs scraped from the Internet Archive.\footnote{\url{https://archive.org/}} 
The articles 
 come from a set of 38 publishers and cover diverse topics. Finally, the Extreme Summarization (XSum) dataset \citep{xsum} consists of $\sim$230k article summary pairs taken from the BBC. \footnote{\url{https://www.bbc.com/}} 
 Each summary is a single sentence long and is professionally written (usually by the author).
For all datasets, we use the splits defined in the original works that proposed them. Because the datasets are too large to provide as supplementary material, we provide pointers in the source code README for acquiring them.

\paragraph{Hyperparameters} Details about important hyperparameters can be found in Section 4 of the paper. Additional training hyperparameters can be found as the default parameters in the training script of the source code.\footnote{\url{https://github.com/Andrew03/transformer-abstractive-summarization}} Most hyperparameter values selected were the same ones suggested by previous work on transformer language models \cite{gpt}. The only hyperparameter we varied that is not measured as an ablation (i.e., training schedules and whether to include source embeddings) was the initialization of source embeddings (if they were included). For this hyperparameter, we explored three different initializations: 1) initializing both source embeddings with zero vectors, 2) initializing both source embeddings with values sampled from the standard normal distribution, and 3) initializing both source embeddings with values sampled from a normal distribution with mean 0 and standard deviation equal to half the norm of the average norm among pretrained embeddings from the GPT language model. This last one is the one we report in all experiments.

\paragraph{Experimental Process} 
Each experiment was run as follows for any given model and dataset. First, we trained the model as described in the paper. After every 1000 minibatches, we compute ROUGE for a random, but persistent 500 example subset of the validation set. When the ROUGE-1 score of the model stopped rising, we used the previous checkpoint as a model to generate summaries for all articles in the test set. We used beam search to decode summaries using a beam with of 3. \textbf{We ran exactly one evaluation run for each result we include in our paper.}

\end{document}